# Accelerating Domain-Aware Electron Microscopy Analysis Using Deep Learning Models with Synthetic Data and Image-Wide Confidence Scoring

**Authors:** ***M.J. Lynch[1*], R. Jacobs[2], G. Bruno[1,] P. Patki[3], D. Morgan[2], K.G. Field[1]***

**1: University of Michigan - Ann Arbor**

**2: University of Wisconsin – Madison**

**3: Former University of Michigan, Currently at Intel Corporation**

***: Corresponding Author**

## Abstract

The integration of machine learning (ML) models enhances the efficiency, affordability, and reliability of feature detection in microscopy, yet their development and applicability are hindered by the dependency on scarce and often flawed manually labeled datasets and a lack of domain awareness. We addressed these challenges by creating a physics-based synthetic image and data generator, resulting in a machine learning model that achieves comparable precision (0.86), recall (0.63), F1 scores (0.71), and engineering property predictions ($R^2$=0.82) to a model trained on human-labeled data. We enhanced both models by using feature prediction confidence scores to derive an image-wide confidence metric, enabling simple thresholding to eliminate ambiguous and out-of-domain images resulting in performance boosts of 5-30% with a filtering-out rate of 25%. Our study demonstrates that synthetic data can eliminate human reliance in ML and provides a means for domain awareness in cases where many feature detections per image are needed.

## 1. Introduction

The development of accurate and practically useful supervised machine learning (ML) models such as deep neural networks (NN) for object detection primarily hinges on the availability of accurately labeled training data. However, obtaining such data is often resource-intensive, requiring considerable time and financial investment. In particular, in microscopy, obtaining a high-quality database of images of widely varying materials and/or imaging conditions can easily require hundreds or more of laboratory hours, that includes sample preparation, sample stimulus (*e.g.*, heating, irradiation, corrosive media exposure, etc.), imaging, and post-processing and labeling. Thus, image set data generation could easily represent multiple person-years with research expenditures exceeding millions of dollars in some fields, *e.g.*, nuclear materials. Furthermore, production of labeled data via traditional human-based methods are susceptible to human bias and error [1], [2], [3], [4], [5], which leads to wide variability within human-derived image datasets. Human-based inconsistencies can only be reduced, but not eliminated by resource-intensive manual reviews. One emerging method to account for the intrinsic challenges with human labeling has been the use of consensus crowdsourcing [6], [7], [8] to build databases. While these procedures excel at using probabilistic models to acquire agreement on features with relatively small levels of ambiguity, challenges remain when analyzing rare or highly ambiguous features. These challenges are exacerbated when using any human-involved system (and especially when performing consensus crowdsourcing of nonexperts), as even experts in a field can disagree on feature presence, precise location, and exact feature size and morphology. The issues around traditionally labeled datasets are only exacerbated with large datasets of a thousand or more features and features within complex, microscopy-derived images. Such datasets often employ multiple labelers, each with subtly different explicit or implicit feature identification criteria and previous labeling experience, resulting in an intrinsic quality variance for these large or complex datasets. Finally, while some features may exhibit similar visual properties (such as dislocation loops and helium bubbles), models trained to identify a single feature type often struggle on a near-similar feature type [9], resulting in limited transferability of datasets between feature types, which leads to the creation of highly specific datasets on a per feature or group of features foundation.

Synthetic data have been widely discussed as a potential method to generate scalable and domain-specific, bias-free datasets and have been shown to have utility in non-electron [10], [11] and electron microscopy settings [12], [13]. 'Synthetic' data generally refers to data that is synthesized, *i.e.*, created by some alternative means that differs from the typical 'natural' source of such data. Here, the synthetic approach is from a physics-based computational model, so we will also refer to synthetic data as 'simulated' and natural data as 'experimental'. In this work, our synthetic data consists of three key components: (a) an experimentally obtained electron microscopy-based phase contrast image with no features of interest present (*i.e.*, 'clean' experimental background images); (b) computationally simulated features derived from physics-based models that have been added to the clean background image, mimicking a natural experimental image feature with known physical parameters; and (c) automatically generated, pixel-precise labels of the added features, eliminating the need for manual labeling and its associated complexities (*e.g.*, human bias). The 'clean' experimental background images can be sourced from literature or manually collected. Formation of our synthetic images takes significantly less time than collecting images containing specific features, as they require no lengthy post-collection processing (*e.g.*, feature labeling), and they can be reused in different synthetic databases.

In addition to addressing the challenge of data generation, this work also addresses the problem of determining a model's domain of applicability for optimal performance. A major challenge for ML-based computer vision (CV) models is the awareness of domain boundaries and identifying when a prediction is being made on features outside the model's realm of applicability. This challenge is not unique to detecting features in microscopy images and occurs in general computer vision and regression- and classification-based ML models [14]. Many convolutional-NNs (such as those used here) are often considered black boxes, with minimal understanding of the specific underlying procedure or physics of the imaged system. This apparent opaqueness creates challenges in identifying model domain limits and hinders assigning a specific cause when model performance is found to be lacking. As there is no common mechanism to tell when new test data is inside or outside a model domain, any procedure that enables establishment of performance guardrails, and thus allows a model to 'self-regulate' within its application domain, would be broadly applicable and highly valuable. This is especially true in scientific and engineering applications where safety factors and other decision points could be derived from ML-based inferences. In this work, we use a post-inference evaluation method based on a model's object prediction confidence to roughly determine our trained models' domain boundaries, providing desirable 'guard rails' that can improve the inference performance and consistency.

Our overall framework is shown in Fig. 1 and integrates SIGMA-ML (Synthetic Image Generation for Microstructural Analysis with Machine Learning), a fully autonomous pipeline capable of generating synthetic, ready-to-train datasets for feature detection in electron microscopy images with an integrated self-regulation post-inference workflow. We then use a conventionally trained NN to produce object detection models (here, we use You Only Look Once (YOLO) model architectures) that are coupled to our self-regulation technique which is shown within the workflow of Fig. 1. The workflow culminates into a final whole image threshold to provide high quality, in-domain model predictions.

We present results using the framework in Fig. 1 for radiation-induced cavities as an example feature type. Cavities were selected for three reasons. First, our team's expertise with the collection and analysis of cavity micrographs significantly aided our evaluation of the quality and realism of the synthetic images. Second, cavities are the central feature of interest in studying swelling in nuclear materials where the size and number density of cavities directly correlates to macroscale volumetric change (*i.e.*, swelling) of a component while in service. Swelling is a key degradation process for nuclear materials and thus a wide body of work using both human-based analysis [15], [16], [17], [18], [19] and CV-based analysis [20], [21], [22] already existed at the time of this study for use in benchmarking our results. Lastly, the relative size and apparent contrast of cavities in transmission electron microscopy images have similar length scales to other distracting features (see Fig. 2 for examples), creating situations of feature ambiguity. With human-based labeling, feature ambiguity results in no universally agreed upon ground truth – as we will demonstrate later.

We show that ML models trained using the SIGMA-ML approach for generation of synthetic bright-field (BF) transmission electron microscopy (TEM) images of irradiation-induced cavities in metal alloys, Fig. 1, outperform or match the performance of models trained on manually labeled, experimentally-derived data of the same domain used in Jacobs *et al.* [21]. We evaluate performance both using traditional and canonical computer vision (CV) metrics like an F1 score (*e.g.*, pixel-based, object-based, etc.) and engineering metrics such as feature size, number density, etc. used in the general physical sciences fields. Additionally, we demonstrate that a simple approach for model self-regulation is achievable based on aggregated single feature confidence scores, providing a powerful means for filtering out-of-domain images and ambiguous feature sets in any ML inference pipeline.

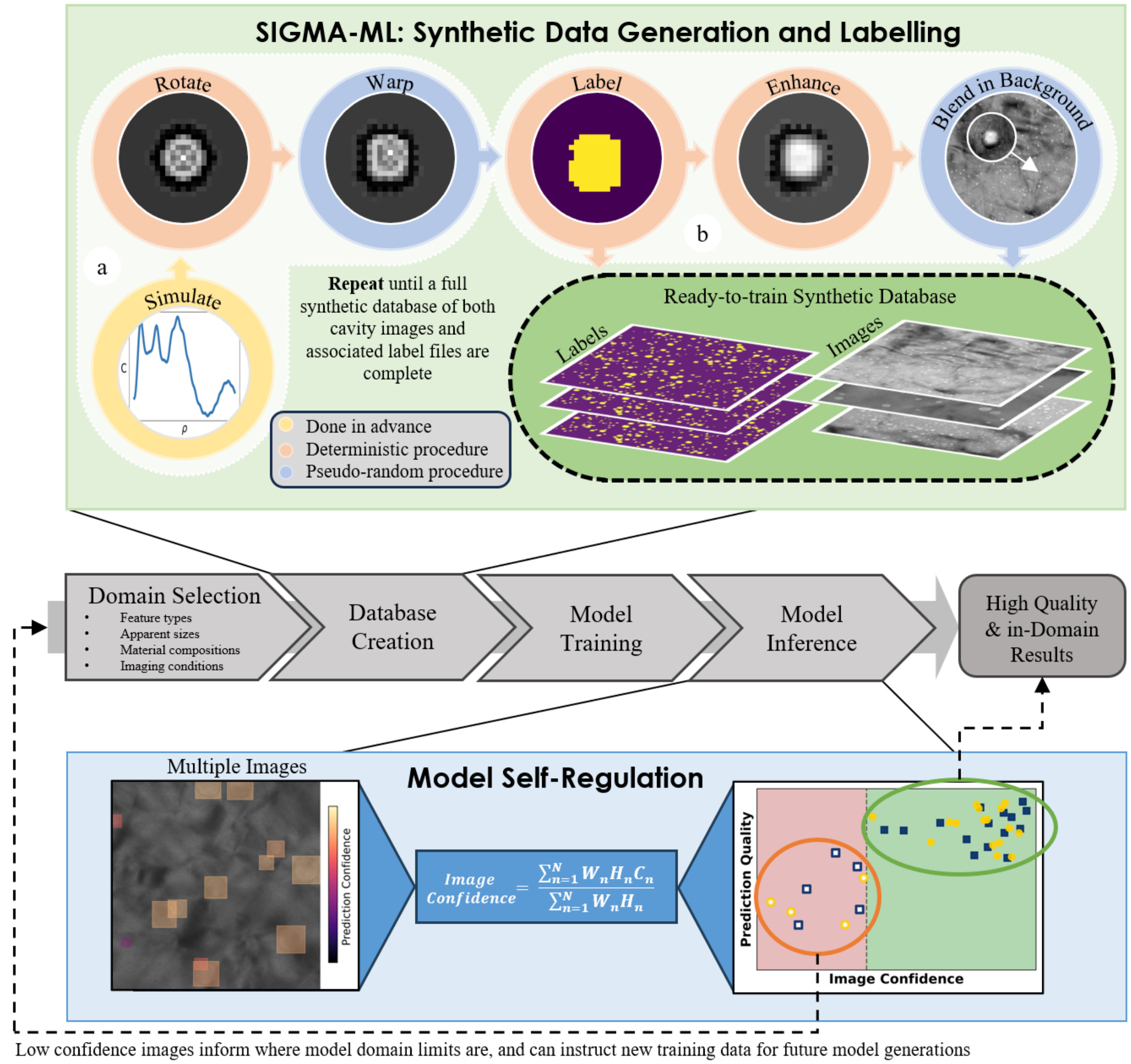

***Figure 1:*** *Overview of our procedure for generating synthetic databases with SIGMA-ML (top – green box). The first few steps (a) are feature-specific and are shown using radiation induced cavities here, however different features of interest could be used with only slight modifications to the initial steps. The remaining steps (b), provide the pixel-precise feature labels and realistic synthetic images used in model training. Our model-self regulation schema (bottom – blue box) shows how image predictions are used to calculate an image confidence score for each*

*image, this correlates well with the quality of the predictions in the image, thus enabling us to flag or remove likely challenging images from inference, improving confidence in model performance.*

## 2. Results & Discussion

**2.1: Comparison of Synthetically and Experimentally trained models.** The performance of our trained Synthetic and Experimental models (see Methods section for details) were both evaluated using the same test set of 20 randomly selected experimental images containing a total of 1108 radiation-induced cavities across all images in the under-focused BF condition. The results of the Experimental and Synthetic models are quantified in Table 1 and select images from the test dataset with model predictions are shown in Fig 2. A detailed description of the experimental database, including its general domain, is provided in Jacobs *et al.* [21]. The work of Jacobs *et al.* [21] also includes our previous best efforts using a Mask R-CNN (R-CNN = Regional Convolutional Neural Network) experimental-ML model for automated cavity detection and quantification of swelling.

Both models exhibited their best performance on the test dataset at approximately the 700$^{th}$ epoch. To ensure this performance is not an anomaly derived from the training process, both models were also evaluated over a range of epochs. This range included epochs 400 to 1000, in increments of 100 epochs. The performance of both models at each of these seven epochs was averaged together to further demonstrate the equivalency in training and final capability of our Experimental and Synthetic training data. The values at both the 700$^{th}$ and averaged epoch conditions are reported in Table 1. As all values of the 700$^{th}$ epoch were found to be within 2$\sigma$ of the averaged epoch conditions, as given in Table 1, the remainder of the discussion will solely focus on the model assessments from the 700th epoch.

The Experimental and Synthetic models exhibited near identical performance with each other across all evaluation metrics, regardless of if our proposed self-regulation technique was used. Both models consistently exhibited higher precision than recall, resulting in a F1-score strongly weighted down by the lower recall performance. However, both models still had overall strong performance for cavity feature detection tasks, with both having an F1 of 0.71 when self-regulation was not applied. An F1 score of 0.71 is comparable to other reported models on similar tasks (discussed later), and effectively identical to the F1 of 0.70 found in our previous work, Jacobs *et al.* [21].

*Table 1:* ***ML Model Results on the Test Set of 20 experimental images.*** *Values are calculated by taking the mean of all image values for a given metric. 'Best' represents the performance of a single weights file. 'Average' is the mean performance of 7 evenly spaced sets of weights between the given epochs. The uncertainty for the "Average" values is the standard deviation of the 7 values that were averaged.*

| **Training Data Domain filtering** | **Selected Weights (Epoch Number)** | **Unfiltered Images** | **CV Metrics** | | | **Engineering Metrics (Swelling)** | |
|---|---|---|---|---|---|---|---|
| | | | P | R | F1 | $R^2$ | RMSE (%) |
| Experimental | Best (700) | 20 | 0.86 | 0.64 | 0.71 | 0.81 | 0.14 |
| | Average (400-1000) | 20 | 0.87 ±.01 | 0.63 ±.02 | 0.70 ±.02 | 0.73 ±.11 | 0.17 ±.05 |
| Synthetic | Best (700) | 20 | 0.86 | 0.63 | 0.71 | 0.82 | 0.14 |
| | Average (400-1000) | 20 | 0.81 ±.03 | 0.62 ±.02 | 0.69 ±.02 | 0.70 ±.15 | 0.18 ±.04 |
| Experimental w/ Self-Regulation | Best (700) | 16 | 0.91 | 0.73 | 0.80 | 0.91 | 0.10 |
| | Average (400-1000) | 17 | 0.90 ±.01 | 0.69 ±.02 | 0.76 ±.02 | 0.87 ±.03 | 0.12 ±.01 |
| Synthetic | Best (700) | 15 | 0.91 | 0.72 | 0.80 | 0.91 | 0.10 |
| w/ Self-Regulation | Average (400-1000) | 15 | 0.87 ±.03 | 0.68 ±.02 | 0.76 ±.02 | 0.84 ±.07 | 0.14 ±.06 |

Self-regulation of the models was applied by using a simple approach to calculate a new quantity we refer to as the overall image confidence score. The overall image confidence score is determined by the size and confidence values of the individual feature predictions for a given image. If an image falls below a given threshold for this metric, the image is flagged as out-of-domain and filtered out, thus removing it from subsequent analysis. An image confidence score threshold of 0.7 was uniformly applied in this work. Images that are filtered out are considered out-of-domain because they contain a high fraction of overly ambiguous and/or non-sensical features to the ML model, resulting in the object detection model making poor predictions. More details on the applied methodology are provided within the Methods section.

Application of our self-regulation method produces two major results. The first result is a noticeable improvement across all models studied in terms of both CV and engineering metrics as shown in Table 1. This improvement underlines the effectiveness of our method in consistently improving model performance on a given dataset. Note, self-regulation had the ability to automatically assess and establish model domain boundaries without access to the ground truth, affirming the utility of our method. The second result of the self-regulation method is that the Synthetic model had a higher average filtering rate of 25% than the Experimental model, with an average filtering rate of 15% (Table 2). A likely cause of this difference in filtering rate is the synthetic data image domain does not have exact agreement with the experimental training set and test image set domain. This disparity can be due to differences in individual defect characteristics (such as real/apparent size or morphology), as well as image-wide characteristics (such as imaging conditions, background contrast, and/or background noise). In our case, the precision of SIGMA-ML enabled the variance in defect characteristics to be minimized, but image-wide characteristics could not be highly controlled due to the random selection of cavity-less background images and the stochastic nature of the background contrast within those images. Here, we assume a leading factor in the difference of the filtering rate is attributed to image-wide characteristics, but this assumption is difficult to accurately test since removal of backgrounds from experimental images would create further complexities and no test sets that fully deconvoluted defect and image-wide characteristics could be created within the scope of the present work. The visualization of the ground truth annotations and the predictions of both models on a high and low performing image from the test set are shown in Fig 2. Both Experimental and Synthetic model predictions for Image 16, Fig. 2 – top, were nearly identical with each other and the ground truth, with an F1 score of 0.92 and 0.90 respectively. Image 18, shown in Fig. 2 – bottom, was one of the worst evaluated images in the test set with F1 scores of 0.51 for the Synthetic model and 0.57 for the Experimental model, however Image 18 was flagged and automatically filtered out by our self-regulation approach for both models. One factor contributing to the poor performance of both models on Image 18 is what we believe to be significant errors in the original ground truth annotations used within Ref. [21], where extensive over-labeling is observed. Note, this issue was only observed after analysis in this work and not our original efforts within Ref. [18]. The role and limitations of human labeling are discussed in Section 2.2.

Ground truth labeling errors cannot be the only cause of poor performance for Image 18 and similar images in the test set because the domain filter functions without any awareness of the ground truth annotations. Another likely factor limiting model performance is the apparent size of features in each image. Previous works have demonstrated the challenges for both humans [1], [2] and NN-based ML models [23] in labeling small features, especially for ambiguous features such as cavities [21], [24]. Therefore, this is a current limitation with existing NN-based ML algorithms, including the YOLOv7 model used here. For example, the test set consists of images with an average apparent feature size ranging from 2.0% to 9.8% of the total image size and a median value of 3.4%. These values are calculated as the mean of the width and height of each prediction bounding box, as a fraction of the total width and height of the image. Image 16 has an average relative feature size of 5.0%, while Image 18 has an average relative feature size of 2.9%. In fact, all the images selected by the domain filter had an average relative feature size below 3.0%, showing that these small-feature-dominated images are exclusively where both models struggle and suggests the self-regulation approach used provides some flagging characteristic for images containing ambiguous and difficult-to-detect features. Care was taken to ensure that both synthetic and experimental training sets had similar distributions of apparent feature size, see the supplemental information (SI.1) for details.

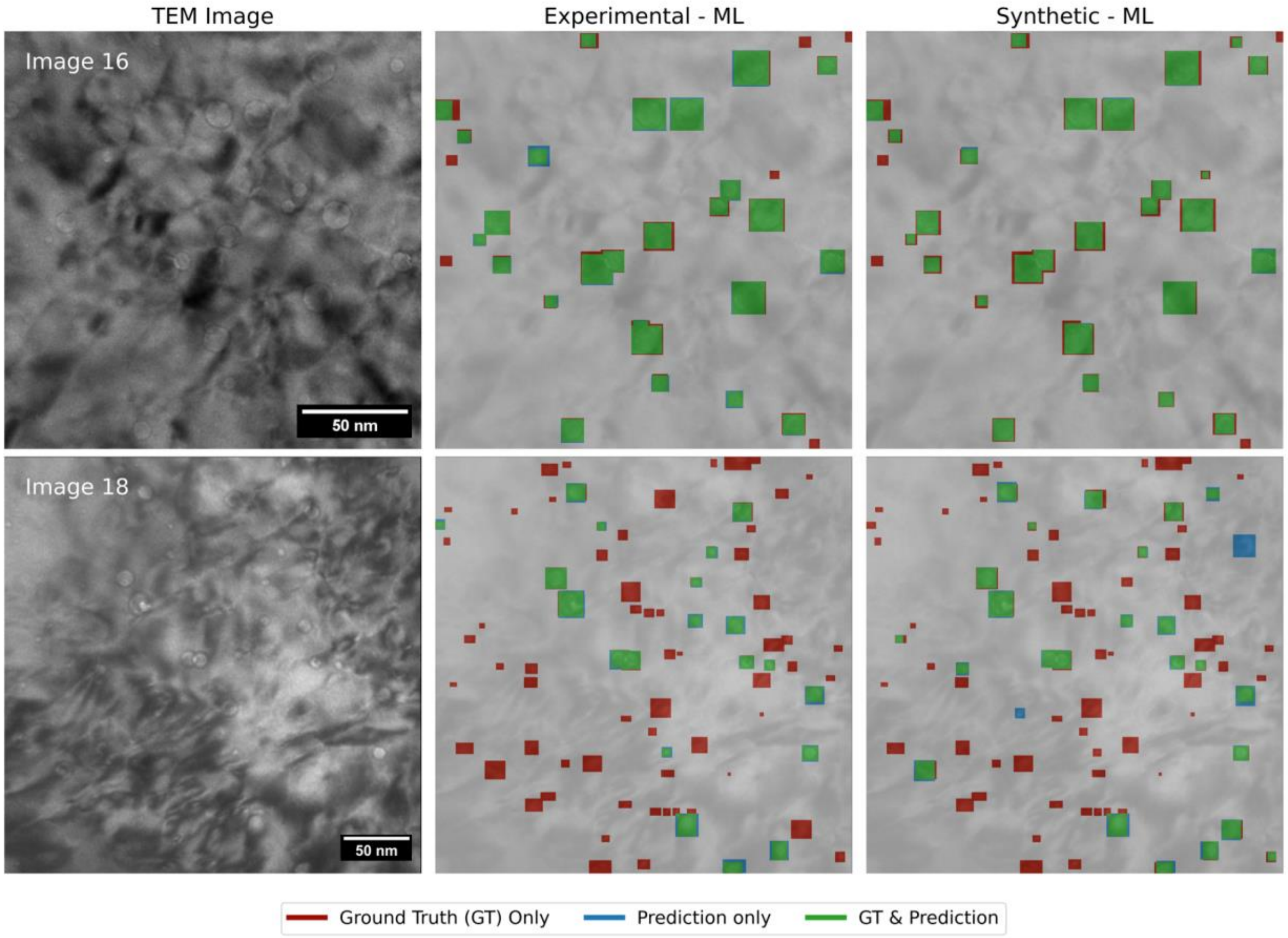

*Figure 2:* ***TEM BF images and prediction visualizations for a high (top) and low (bottom) performing image.*** *Image 16 (top) & 18 (bottom) from the test dataset. The Experimental-ML model had a pixel-wise F1 score on Image 16 (top) of 0.92 while the Synthetic-ML model had a score of 0.90. On Image 18 (bottom) the models had pixel-wise F1 scores of 0.51 (Experimental) and 0.57 (Synthetic). In both images, models have similar performance variance characteristics for evaluated engineering performance metrics. Image 18 also shows an author-perceived over labeling issue present in the datasets obtained from* [21].

To inspect the influence of our test dataset selection, we evaluated our Synthetic model on the full Experimental dataset, which contains 13,956 under-focused cavities from 175 experimental images available in this study. This could not be done with our Experimental model as many of these images were used in training the Experimental model. Testing on this expanded dataset with our 700 epoch Synthetic model with self-regulation achieved an F1 score of 0.71 and an image filtering-out rate of ~40% was found, where without self-regulation a lower F1 of 0.62 was achieved. While still reasonably close to the model performance on the Test dataset with an F1 score of 0.80 with self-regulation, this 11% decrease in F1 is primarily caused by a significant decrease of Recall from 0.72 to 0.63. We hypothesize the drop in performance is likely an effect of errors in the ground truth labels, discussed in Section 2.2. Our hypothesis is supported by the observation that enhancing model selectivity – by modifying feature and image confidence thresholds, as detailed in Table 2 – resulted in minimal changes on model performance.

**2.2 Quantification of human performance and assessment against ML models.** The results in the previous section suggested that overall performance could be impacted by the human-based selection of features comprising the ground truth labels. To evaluate the performance of our models against human-based labeling, a labeling survey among six human experts and both ML models on an additional five test images not previously used were performed – details on how this was conducted are in the Methods section. The results are shown in Fig. 3, where a normalized swelling value is used to better visualize deviations across the labeling survey image set. From Fig. 3a, experts often deviated from the designated ground truth, but more importantly, our models consistently performed within these deviations, making them indistinguishable from human expert labeling if the *a priori* knowledge of who/what performed the labeling was unknown.

The statistically indistinguishable nature of our models to human labeling is further supported in Fig. 3b, where the normalized swelling distribution of the twenty test images and five labeling survey images shows that all three methods had very similar mean absolute error of predicted normalized swelling of 16%, 21%, and 21% for the Experimental, Synthetic, and human experts, respectively. The deviations in Fig 3a, which extend into Fig 3b, are largely driven by images dominated by small features. For example, Image B & E proved particularly challenging for the experts to have consensus on, with a standard deviation in normalized swelling of 0.17 and 0.44, respectively, while Image A had higher consensus with a much smaller standard deviation of 0.08. This finding reflects the inherent challenges of consistently labeling images with fine details. Nevertheless, the tendency of ML models to underpredict normalized swelling was consistent, potentially attributed to the previously discussed reasons of errors in the ground truth and known challenges in small object detection for CV models. Importantly, the underprediction of swelling of the ML models shown in Fig 3a remained within the inherent scatter of human labeling efforts on the same Image, again suggesting indistinguishable performance for the Synthetic model to either the Experiment model as shown in Table 1 or human-based labeling methods.

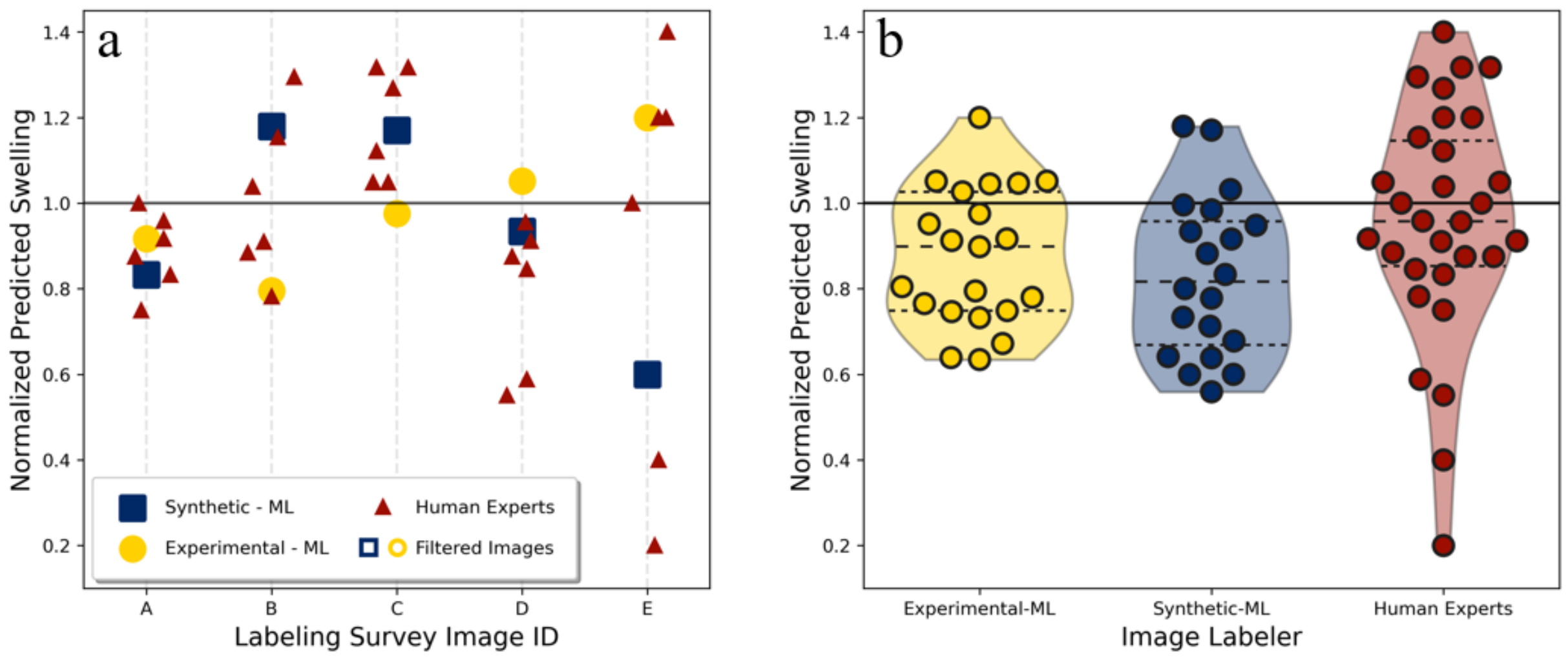

***Figure 3: ML models predict image swelling within the range of human performance.*** *(a) Per image results from the labeling survey that demonstrate both ML models consistently predict image swelling within the range generated by six human experts. Horizontal jitter of Expert results (red triangles) is artificial, and to prevent data overlap. (b) Distributions of the predicted normalized swelling values from the human expert labeling survey, and both ML models on the Test set & Survey set. Image confidence filtering was used for both ML models. Experimental-ML: n =21, Synthetic-ML: n = 20, Human Experts: n = 30 (6 experts x 5 images each = 30).*

The findings of the labeling survey (Fig 3a) were further validated using each of the six human expert label sets as the acting ground truth. These seven iterations of analysis found an expected minimum mean error of ~20%, see SI.2 for details. This is consistent with the collective experiences of the authors regarding challenges reaching a precise and universally agreed-upon swelling value. Findings were also reviewed as a function of absolute swelling and no significant correlation was found, *e.g.*, high/low swelling images did not have more or less expected error in predicted swelling.

**2.3: The effectiveness of self-regulation through image confidence score filtering.** Fig 4 shows the impact of our self-regulation approach using our image confidence score filter method. See SI.3 for examples of the prediction size-confidence relationship and calculation of image-wide confidence. A notable correlation exists between the image confidence score and the computed F1 for both model types. The optimum value for an image confidence score threshold is one that simultaneously maximizes prediction performance and image inclusivity. This balance was found to be at an image confidence score of 0.7. This value is not universally applicable to all model scenarios and in this case was selected after an in-depth sensitivity analysis was completed, as detailed in SI.4. Future uses of this approach would likely need to undergo a similar sensitivity analysis to best calibrate both thresholds for a specific application. This customization enables users to decide how selective they want their analysis to be, for example, decreasing the object-wise prediction threshold while increasing the image-wise threshold generally results in increasing prediction performance but decreasing image inclusivity. From carefully observing the excluded images, we believe that this method for identifying out-of-domain images identifies images that are both out of the scientific imaging domain of the model (*e.g.*, different features, imaging conditions, *etc.*) and images that contain highly ambiguous or complex features due to either background noise, poor image quality, or both. The correlation between F1 and image confidence substantiates our ability to use the image confidence score as a general model performance predictor, thereby aiding in defining model domain boundaries. Similar results as shown in Fig. 4 were observed when comparing image confidence with the previously discussed engineering evaluation metric (material swelling), see SI.5 for details.

Efforts were made to utilize published cavity data from Chen et al. [24] to further evaluate the domain-awareness of the image confidence score filtering method on generally in-domain images. A portion of the Chen data included cropped sections of larger TEM cavity images, drastically increasing the apparent diameter of the cavities to 10-30% of the image size. As our model trained on synthetic data was trained on standard full-size images, where features are <10% of the image size, this difference in feature size created a significant mismatch between these two image domains. Our image confidence score filter was successful in removing most of the Chen images containing large apparent cavities, further demonstrating its utility. See SI.6 for a detailed comparison of database feature sizes. Note, due to the test dataset used in Chen et al. [24] being unavailable to the authors, a direct comparison of our models against their model CV and engineering metric performance was not possible.

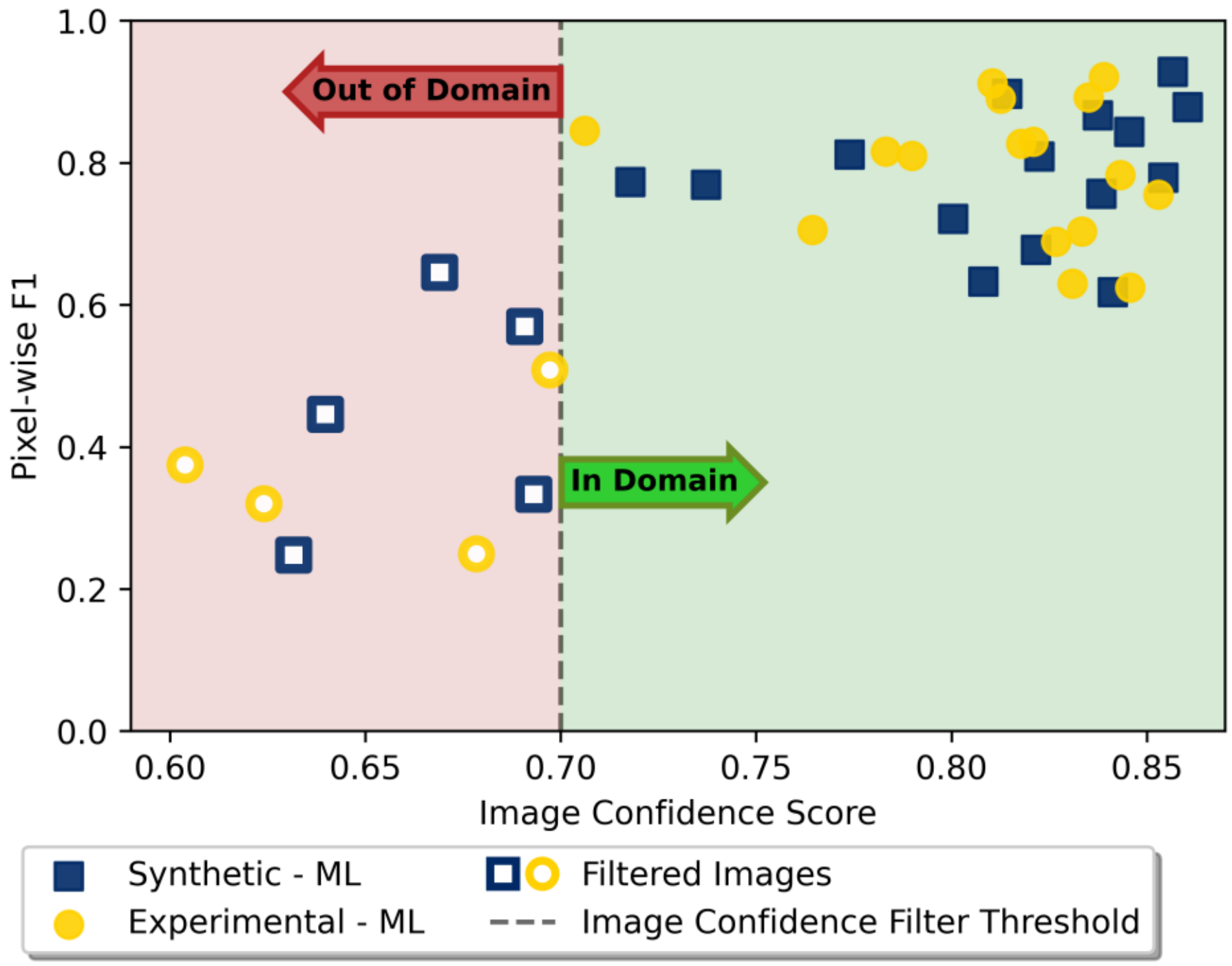

***Figure 4: Image confidence score is a good indicator of F1 performance.*** *Both models were evaluated using the Test image set (n=20) and had the same self-regulation process applied. Individual points represent the image confidence (calculated using the predictions of the model), and the calculated F1 score.*

We further validated the application of our self-regulation approach by performing inference and evaluating the model filtering rate on increasingly out-of-domain image sets, summarized in Table 2. A dataset of electron microscopy images of radiation-induced dislocation loops from Jacobs et al. [25] was used as an example of images that are 'near domain' and for which we here refer to as the 'Loops' dataset. While these images were collected under different imaging conditions (on-zone bright field scanning transmission electron microscopy) and are of different feature types, they still exhibit somewhat similar visual characteristics with our cavity test dataset, such as general background appearance/noise and approximately round features with a dark outer ring. Similarity is expected in this case as transmission electron microscopy and scanning transmission electron microscopy are also reciprocal images when properly configured, providing images of similar visual parity [26], [27]. Nevertheless, our Synthetic model flagged 60% of these images as out of domain using the same thresholds as with our cavity inference. With only slight adjustments of the threshold (0.35 for individual features, and 0.75 for images), this filtering rate rose to 91% on this Loops dataset. The same adjusted thresholds only changed the filtering rate on the Cavity Test Set from 25% to 35%.

A 128-image subset [28] of the larger COCO [29] database of common objects was used as an example of a completely 'out-of-domain' test set to evaluate how self-regulation worked on non-electron microscopy images that exhibited minimal similarities to the trained feature type or imaging conditions. A filtering rate of 88% was found using the standard threshold values, and 95% when using the more selective values. Of the few unfiltered images, common traits such as large round features with significant contrast change (*e.g.*, large white letters on a black background) were often found. These features roughly mimic common visual characteristics of cavities and had an average size of 9.2% of the total image. These two factors lead to high individual feature confidence scores, and with our feature size weighting on image confidence, a correspondingly high image score resulting in being above the set threshold marks. Table 2 further shows capability of domain awareness, as when even image inclusivity was set to be maximal, the two more similar datasets (Expanded Cavity & Loops) had minimal filtering rates, while the COCO 'out-of-domain' set still had a significant filtering rate of 72%. The variance based on the adjusted thresholds highlights how this method can be tailored for a project or specific application, preferring either image inclusivity or confidence in prediction performance.

*Table 2: **Quantification of Self-regulation on In-Domain and Out-of-Domain Datasets.** The highly sensitive nature of both filtering thresholds is apparent by the significant changes in filtering rates with only minor threshold adjustments. The same Synthetic model was used in all evaluations.*

| **Image Inclusivity** | High | Standard | Low |
|---|---|---|---|

| Individual Prediction & Image Confidence Score Threshold | | 0.45 & 0.65 | 0.4 & 0.7 | 0.35 & 0.75 |
|---|---|---|---|---|
| **Inference Set** | **Image Count** | **Image Filtering rate with varied thresholds** | | |
| ***Cavity Test Set*** | 20 | 0% | 25% | 35% |
| ***Expanded Cavity Set*** | 175 | 16% | 41% | 60% |
| ***Loops Set*** [25] | 107 | 24% | 60% | 91% |
| ***COCO 128*** [28] | 128 | 72% | 88% | 95% |

As absolute swelling can vary significantly between images, a normalized swelling value is useful for comparison of performance between images and can further provide insight on self-regulation. Fig 5a compares the predicted normalized swelling value of each image across the Cavity Test Set. There is high agreement between the two models, when a predicted image passes both filters, with a median difference between the Synthetic & Experimental model being 0.086 (normalized units). Both models tend to underpredict the image swelling values with respect to the ground truth, with an average difference from the normalized predicted swelling of 0.18 and 0.21, respectively for the Experimental and Synthetic models when using the image domain filter method. This difference is likely caused by two effects, the first being the models underestimating the number of features and the second being the previously mentioned issues with over labeling in the ground truth, as demonstrated for Image 18 in Fig. 2. The under-labeling of features could be due to either errors in its training data labels or a slight domain mismatch between the synthetic images generated by SIGMA-ML and the test domain. It is believed the former is more relevant in the case of the Experimental model whereas the latter explanation is more likely impacting the Synthetic model. The ground truth over labeling issue likely has a higher impact, as revealed by the Recall scores of 0.73 and 0.72 for the Experimental and Synthetic models in Table 1 indicating that both models are not finding all ground truth features that would be used within the normalized swelling calculations for Fig. 5.

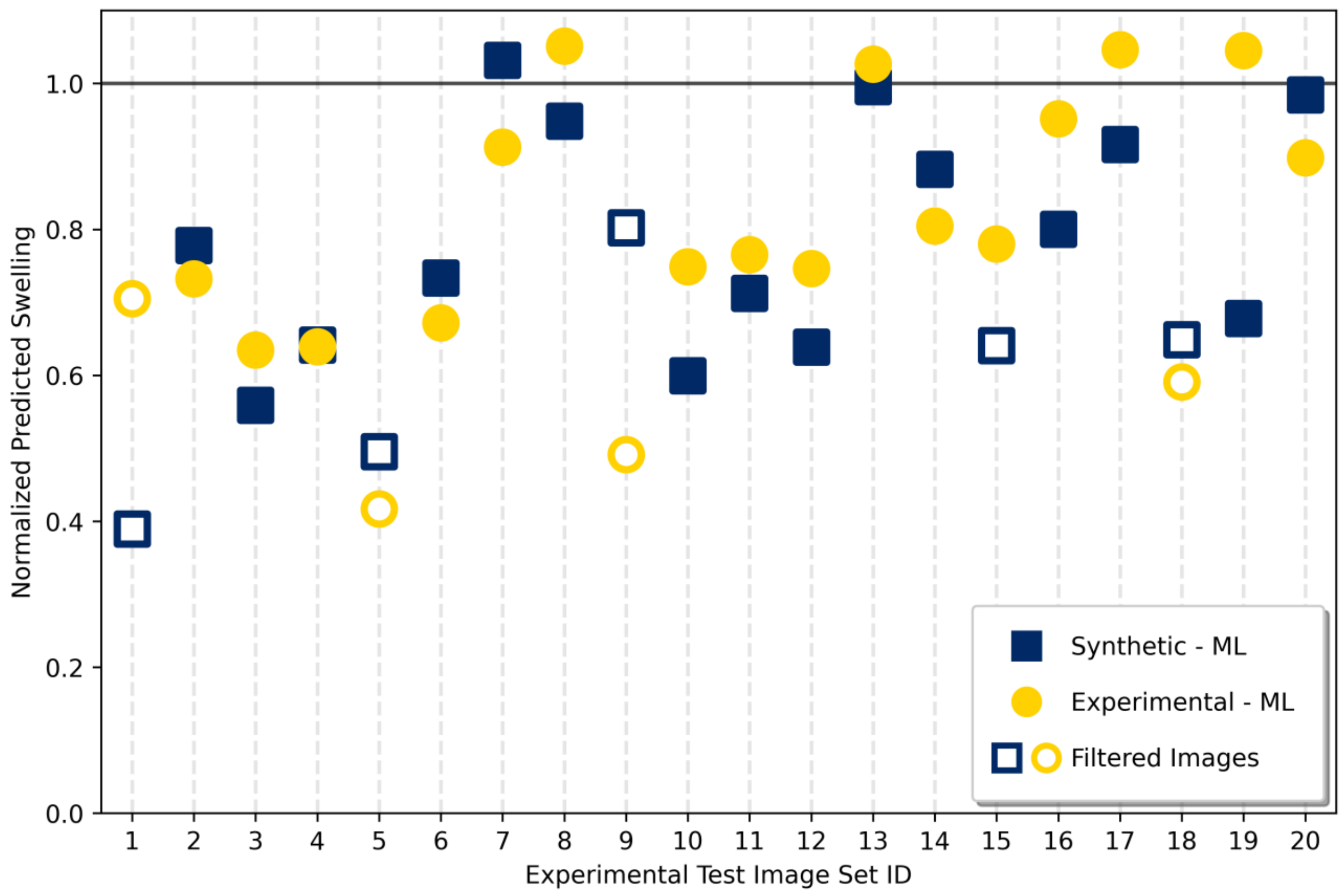

*Figure 5:* ***Normalized predicted swelling values from ML models & Experts.*** *Each image ground truth swelling is set to be 1.0, unfilled symbols denote images that fail the image confidence score filtering. Predictions are on the 20 image Cavity Test Set.*

**2.4 Modularity of SIGMA-ML with Self-Regulation.** We have demonstrated the effectiveness of synthetic data produced by SIGMA-ML and the utility of our self-regulatory model procedure for the example of cavity defects in electron microscopy images, suggesting their potential utility to a wide range of microstructural defects. As depicted in Fig. 1, these subsystems within the overarching code base are designed as modules, which can be individually adapted with minimal effort to meet specific use-case scenarios. For instance, Fig 1a highlights the initial three steps involved in the creation of cavity-specific synthetic data. With a minor modification, like integrating a different physics-based model such as multi-slice simulations for more complex defect-contrast image formation [30] or other image contrast models [31], additional synthetically generated features or datasets can be produced and then fed into the remaining SIGMA-ML pipeline. Similarly, our self-regulation procedure is a novel but simple approach that could be easily adapted to other ML approaches in electron microscopy and beyond. Only individual prediction coordinates and confidence values are needed, which are very common in most machine vision workflows. As we found model performance improvement of 5%-30% on various evaluation criteria (regardless of model training data), this suggests similar boosts in performance could be gained on previously reported models with minimal effort.

Moreover, the flexibility of SIGMA-ML synergizes with our self-regulating procedure, enabling a “smart” and iterative approach to synthetic model development (Fig 1.), similar to the “ensemble learning-iterative training” (ELIT) method that has been previously discussed [32]. This process would begin by compiling a diverse experimental test dataset, with high-quality human labels. Then, an initial standard synthetic dataset is generated, and model training is executed. The performance of the synthetic model on the high-quality experimental test set is then categorized into two main bins, high prediction quality (in-domain images), and low prediction quality (out-of-domain images). Human experts then review the out-of-domain images for key criteria (feature sizes, background contrast, *etc.*) and use these domain descriptors to inform SIGMA-ML in generating updated synthetic training datasets. This ensures abundant data in areas where the model showed domain uncertainty. The cyclical process could be done iteratively, until model performance is saturated.

**2.5 Evaluation of synthetic data workflows against published ML models**. In this section, we compare our results with other studies of deep learning-based microstructural feature detections from the literature [21], [24], [33], [34], [35]. As shown in Fig. 6, it is evident that both our Experimental and Synthetic model exhibit F1 performances similar to those of other experimentally trained models, and that all ML models are performing near, or above average human expert performance range as determined by our human labeling survey. Analysis of the survey results determined a mean Expert F1 of 0.75 with a standard deviation of 0.13. Dashed lines in Fig. 6 show this range of average performance, wherein 70% of all Expert F1 scores lie.

The models reported herein demonstrated significantly greater (+37%) Precision than Recall, while most previously reported models have these two metrics much closer in value. The use of our domain filter had a larger impact on improving Recall (+14%) compared to Precision (+6%) but a significant difference (+25%) between the two terms still exists. One cause of this imbalance in Precision and Recall is the aforementioned issues with the ground truth, where over-labeling can produce erroneous false negatives which only impact Recall. Even with perfect ground truth files, models would likely still exhibit issues with detecting small features. It is likely that future developments will improve this small feature detection problem. For example, one method to overcome this issue of inability to detect small features was given in Chen et al. [24], where a novel image rescaling technique was used. However, this method is not always applicable, and issues were still found when features were <5 pixels in size (<1% image size). Here, we focused on demonstrating the utility of our synthetic data, rather than the pursuit of maximum model performance.

It should be noted that the model type/architecture, testing set, and training parameters varies widely between all results shown in Fig. 6, so this direct comparison is very qualitative. For example, the highest performing model (Roberts [34]), has an architecture inspired by U-Net [36] and DenseNet [37], and the reported values are from predictions on a test set of a single image cropped out of a training set image. Also, some of these findings are for labeling tasks on other defect structures (faceted voids, loops, black dots). The success of ML methods across these many defect types suggests that future work to expand the capabilities of synthetic defect structures to include these other defects will also yield successful models. Despite these many details we wish to stress that our models, and importantly the Synthetic model, appear to be essentially equivalent to other similar works and expert performance when comparing overall F1 score and engineering metrics alone.

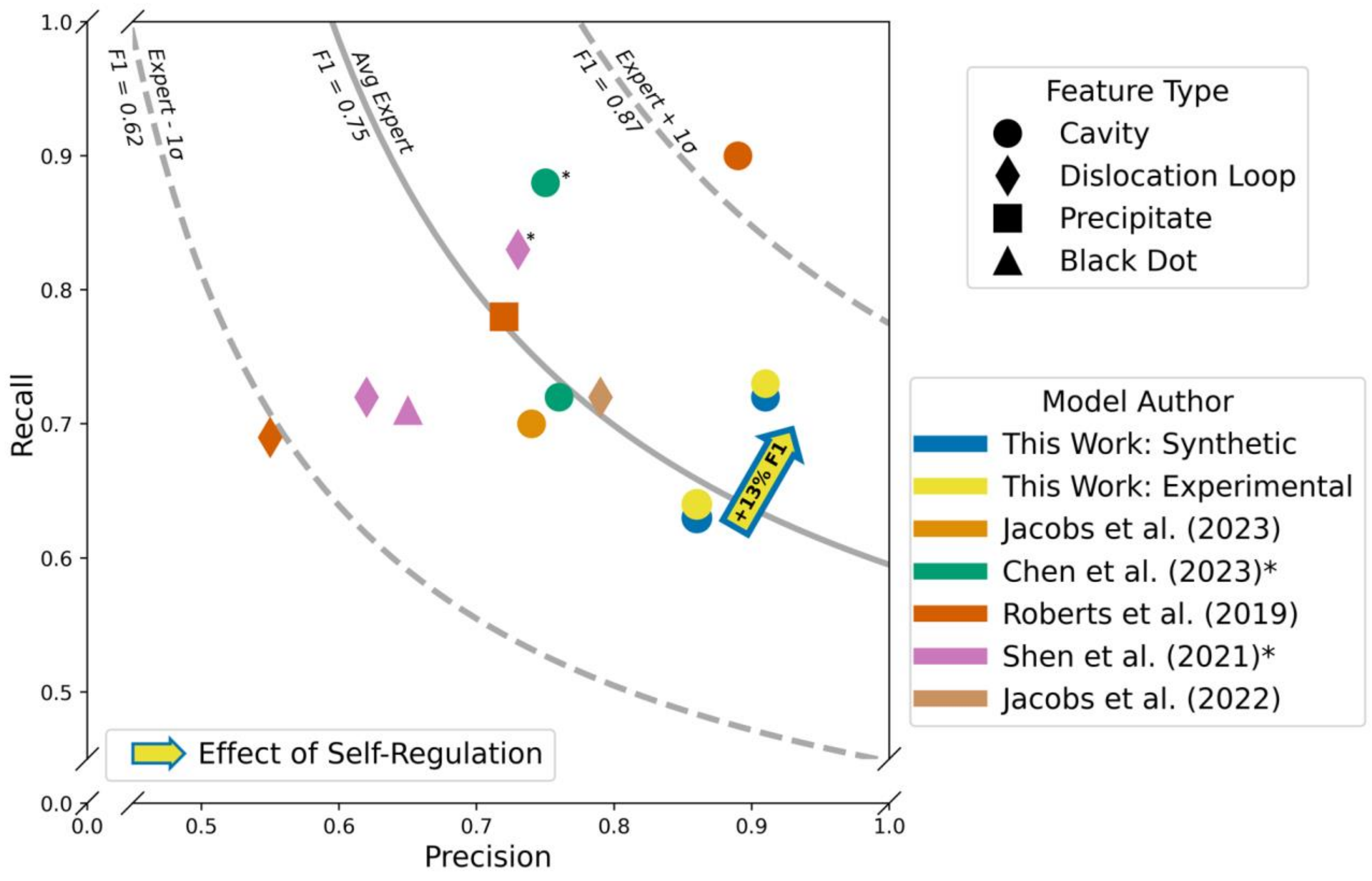

*Figure 6:* ***Comparison of different model performances in irradiated materials electron microscopy.*** *Our findings reported here are displayed with those reported in literature on various different defect structures (cavities, loops, blacks dots), with varying model types, training parameters, and evaluation techniques. Expert bounds are calculated from our round robin results. Helps differentiate between visually identical data. For Chen* [24]*, * denotes the performance using the image rescaling technique. For Shen* [35]*, * denotes loops of the type a/2⟨111⟩, while the other is loops of the type a⟨100⟩.*

**Summary and Conclusions.**

In summary, we have provided substantial evidence suggesting that a synthetic data derived model trained with the YOLO object detection framework demonstrates an almost identical statistical performance as an experimentally derived YOLO-based model across all evaluated metrics for a test case within the field of electron microscopy. Moreover, the introduction of self-regulation via image confidence score filtering led to an enhanced performance observed in all models, emphasizing the power of this simple and flexible method. The schema presented within demonstrates the potential for integrating domain awareness and tailored synthetic data generation by SIGMA-ML to address new and/or out-of-domain gaps with ML-based electron microscopy object detection and quantification tasks. The disparity in the filtering rates when using self-regulation, where the Synthetic model had a higher filtering rate, could be associated with variations among synthetic training data domains, experimental data, and the testing data domain used within this study. This evidence provides a general conclusion that further optimization could be completed either through revision of the experimental data used within, such as culling spurious ground truth labels, and/or through expansion of the synthetic image domain. Domain expansion using SIGMA-ML is only restricted by governing physical equations used for geometry of the cavity and the computational time to generate an increased simulated dataset.

A significant correlation was observed between image confidence score and the calculated F1 for both the Synthetic and Experimental models, highlighting the capacity of an image confidence score as a predictive measure of model performance and means to integrate self-regulation. This lessens the complexity of defining model domain boundaries. Additionally, in comparison to human experts, the ML models maintained a competitive performance level in terms of normalized swelling prediction – the key engineering metric reference in this work, although both tended to underpredict the swelling values compared to the ground truth derived swelling values. This alignment in the performance of both models, coupled with the elimination of human-bias within the inference steps, further strengthens the validity of using ML models for this type of microstructural analysis, even though challenges in labeling small features and errors in ground truth annotations is a remaining limitation for their performance. Within the area of defect detection, our comprehensive procedure effectively confronts the critical challenges around training data quality and domain boundary identification/disparity that are intrinsic to machine learning models. In doing so, our approach unlocks the potential for creating ML models for defect detection that are substantially easier to generate, devoid of human bias, and domain targeted, all while exhibiting performance at or above human levels.

## 3. Methods

**Collection and Labeling of Experimental-ML Data.** Experimental BF TEM images of irradiation-induced cavities in metal alloys were obtained from a variety of sources including the authors, their collaborators, and professional colleagues. A detailed description of the database, including previous work implementing a Mask R-CNN Experimental-ML model, is provided in Jacobs *et al.* [21]. Here, details are reproduced in brevity for completeness. The database of 300+ images were labeled by a domain expert and subsequently corrected by a second expert; this two-step process established the ground truth for all our experimental data. The database contains a wide range of materials such as austenitic and ferritic steels irradiated using both ions and neutrons, additional details on the data domains are provided in our previous work [21]. In this work, we exclusively focus on under focused cavity images within the dataset as that is the most common imaging domain in experimental work and eliminates unwanted complexity in the results due to class imbalances in the training data. A division of all datasets is given in Table 3.

*Table 3:* ***Divisions of different datasets.*** *"Data Type" indicates whether the images comprising a given set are experimental or synthetic images, and "Set Name" represents a dataset naming shorthand for convenient referencing throughout this work. For example, "Test" refers to the test dataset consisting of 20 experimental images which is used to evaluate both the Experimental and Synthetic models. *Validation sets were not ultimately used in any of the models report herein, see Section 'Training of ML models' for details*

| Data Type | Set Name | # of Images (# of Features) |
|---|---|---|
| **Experimental** | **All Experimental Under Focused Images** | **175 (13956)** |
| | Experimental Training | 100 (8441) |
| | Validation | 50 (3998) * |
| | Test | 20 (1108) |
| | Human Round Robin | 5 (409) |
| **Synthetic** | Synthetic Training | 37 (8826) |

**Generation of Synthetic Data.** Experimental TEM images of unirradiated microstructures were obtained from the sources described above and thus encompass the same range of material systems as the irradiated materials database. These images serve as our "clean backgrounds" for the generation of synthetic data. A synthetic database of BF TEM images, containing underfocused cavities was generated using a novel multi-step procedure (Fig. 1, top), termed SIGMA-ML. The SIGMA-ML procedure leverages the simplified defocused contrast theory of Rűhle and Wilkens [38] and mathematical implementation of Foreman et al. [39] to generate the final synthetic images. The full procedure is outlined in Fig. 1 and summarized below.

1.) **Simulate:** The 1D contrast variation from the center to the out radius of cavity as [39]:

$$\psi_n(t+\zeta,\rho) = \psi_{n,p}\left[1 - \frac{2i}{\beta} exp\left[i\frac{\rho^2}{\beta}\right] \int_0^1 \Delta_n(\rho')\, J_0\left(\frac{2\rho\rho'}{\beta}\right) exp\left[i\frac{\rho'^2}{\beta}\right]\rho'\, d\rho'\right] \tag{1}$$

where $\psi$ represent the wavefunction of the electron, $\beta$ is a quantity relating the defocused condition, electron wavevector, and physical cavity radius, $\rho$ is reduced radial distance, $\Delta$ is a complex quantity relating the size/geometry of the cavity and material properties, and $J_0$ is the Bessel function of the first order. Using Eqn. 1, thousands of smooth 1D radial contrast profiles of idealized spherical cavity geometries are simulated for features ranging from 1 – 50 nm radius, and ± 300 – 2300 µm defocus. Typical computation time on a consumer grade laptop (era 2022, 4-core 2.8 GHz) was 12 seconds per simulation. This step is done in advance and simulation results are stored as a lookup table as all features of a given size and defocus in the same material will start with the same initial simulated contrast profile. Identical material and electron beam parameters are used for all simulations. By slightly varying the defocus of individual cavities in the same image, varying degrees of cavity depth in the sample foil are simulated.

2.) **Rotate:** An algorithmically generated list of cavity size/defocus distributions is generated for a new synthetic image and the needed 1D contrast profiles from the previous step are loaded. One contrast profile is selected and then symmetrically rotated 360° around its center, forming a 2D array that can be directly translated to a grayscale image at the appropriate resolution, where the resolution is defined by the background image.

3.) **Warp:** The cavity is then classified as one of three sizes (small, medium, or large) based off the physical radius used in the simulation of step 1. Its symmetrical geometry is randomly warped in all directions based off its size, where large cavities have a higher potential degree of warping. Care was taken to ensure key cavity features, such as the first Fresnel fringe, still appear after warping.

4.) **Label:** Using a modified watershed algorithm, a per-pixel label mask is generated. The edge of the cavity label in the mask is set to be in the center of the darkest outer fringe of the cavity. Rigorous testing on individual synthetic defects showed ~99% success rate in full pixel accuracy of these labels using high resolution visual inspection and mathematical analysis of pixel intensity values. For the ~1% of mislabeled features, they are typically very small features (< ~8 pixel radius) where there are not enough pixels to draw a detailed mask, in this instance the labeling algorithm defaults to square labels ~1.5× larger than the true cavity size.

5.) **Enhance:** Higher-order transmitted electron beam-detector (CCD) physical considerations are then modeled, such as the detector quantum efficiency (DQE) informed by the findings in McMullan et al. [40] and a modular transform functions (MTF) modeled by Van den Broek et al. [41] to help improve physical fidelity by increasing the common stochastic noise apparent in experimental images directly within the simulated cavity image.

6.) **Blend:** Each single cavity is blended into the "clean background" image using a normalization technique with a specific order of operations and the seamlessClone function in the OpenCV2 Python package [42]. The process is repeated for each additional feature until a single image is populated with the targeted density and size(s) of cavities. The image scale (nm/pixel) of the "clean background" image is used to ensure cavities are appropriately sized with respect to background features. Note, in this initial study the overlap of cavities is not considered in our synthetic images due to the observed rarity in the experimental database and the additional complexities in simulated cavity stacks [43] The size distribution and density of synthetic cavities was chosen to match, on average, that of the experimental image database from Jacobs et al. [21]. Images ranged from 0.079 to 0.11 nm/pixel in scale, and 1024×1024 to 4096×4096 pixels in absolute size, note all images are rescaled to the same size during ML model training.

**Training of ML models.** Both model types (Experimental and Synthetic) were trained with minimal augmentation using 'Version 7' of the You Only Look Once (YOLO) algorithm [44] which is derived from the well-established Version 5 of YOLO curated by Ultralytics [45]. The training datasets for both models were developed to approximate the same number of cavities within each dataset with similar pixel-size distributions. Models' performances after training were evaluated using both traditional ML metrics such as pixel-wise Precision, Recall, and F1-score, as well as more experimentally relevant, defect-specific metrics such as percent swelling in our example case of radiation-induced cavities. As noted in Table 3, the synthetic training data consisted of ~3× fewer images than the experimental training set. This reduction in synthetic image count is due to many experimental images having a very low feature density (< 25 per image). Preliminary testing showed changing the feature density in the synthetic dataset and thus the image count had no statistically significant impact on model performance as evaluated using common ML metrics. However, fewer total images resulted in faster training times using YOLOv7. Identical training parameters were used throughout this work and a constant image resolution of 1280×1280 pixels was selected to optimize model detection performance in accordance with the YOLOv7 documentation [44]. The constant image resolution of 1280×1280 pixels results in YOLOv7 up/downscaling most images and using letter boxing methods for non-square images. Trainings were done without a cross-validation dataset. Instead, they were trained to 4000 epochs each, saving weights every 100 epochs. This overtraining is far beyond typical for YOLO models but was done to ensure no local maxima in performance were found and thus reported. Every 100 epoch was benchmarked against the test dataset, and both model types were found to have peak performance at epoch 700. The average of 7 evenly spaced epoch performances is also reported to demonstrate that any model trained to approximately the same range (700 ±300 epochs) would have similar performance.

All trainings were done on an in-house graphical processing unit (GPU) cluster running Ubuntu 20.04 with 4 NVIDIA RTX A5000 GPUs. No transfer learning was used, and the large p6 architecture of YOLOv7 [19] was used, and the only image augmentation operation used was mosaic (no "mixup", as this results in blending of images and overlapping of features), the native YOLOv7 augmentation. Limited augmentation was applied due to the feature per image disparity between the experimental and synthetically generated image sets.

**Performance evaluation.** Here, ML models and human performance is evaluated using classification-centric and materials property-centric metrics. Our key classification-centric metrics are Precision, Recall, and F1 score. For our materials property-centric metrics, we focus primarily on calculated image swelling, as it is a single term that weighs both cavity density and size distributions. Here, we report pixel-wise metrics, as opposed to object-wise, as done in [34]. However, the same overall trends are observed with both methods. Evaluation masks are made for each image in which every pixel is assigned one of four categories, true positive (TP), true negative (TN), false positive (FP), and false negative (FN). The total count of each category is summed, and the classification metrics are calculated with the following equations:

$$Precision\ (P) = \frac{TP}{TP + FP} \tag{2}$$

$$Recall\ (R) = \frac{TP}{TP + FN} \tag{3}$$

$$F1 = \frac{2 * Precision * Recall}{Precision + Recall} \tag{4}$$

Precision (P) is the proportion of predicted feature pixels that are correct. Recall (R) is the proportion of feature pixels in the ground truth that were correctly predicted. F1 is the harmonic mean of P and R, and therefore is weighed more by the lower metric. Perfect labeling with respect to the ground truth occurs when P=R=F1=1. Model inference was done at the default YOLOv7 image resolution of 640x640 pixels, with YOLOv7 handling all image rescaling. This resolution was found to have optimal performance for all model tests, likely due to minimizing image rescaling effects when having to upscale images.

Our materials property metric, image swelling, is given in Eqn. 5. In Eqn. 5, $S$ is the physical area of the image, $\delta$ is the sample thickness, $d_n$ is the cavity diameter which is taken as the average between perpendicular side lengths of the object detection box, and N is the total number of cavities in the image. As TEM sample thickness is often inaccurate or not recorded, we use a constant thickness of 100 nm in all of our swelling calculations which is aligned with our previous calculations [21]. The choice of a constant image thickness means our swelling calculations are areal swelling, which has no impact on the comparison of our results with each other as the calculation still maintains the size and density dependencies for swelling.

$$\frac{\Delta V}{V} = 100 \times \frac{\frac{\pi}{6}\sum_{n=1}^{N} d_n^3}{S\delta - \frac{\pi}{6}\sum_{n=1}^{N} d_n^3} \tag{5}$$

Once evaluation metrics are found they are compared with those calculated from the ground truth using standard error analysis techniques, $R^2$ (coefficient of determination) and root mean squared error (RMSE). Equations for these terms are widely available and not produced within.

**Self-Regulation via Image Domain Filtering.** Individual YOLO model predictions contain five values, where the first four give the location and size of the bounding box, and the last predicted value is object prediction confidence (sometimes called objectness), which ranges from 0 to 1 (low to high). The object confidence is calculated from a linear combination of the conditional class probabilities $\Pr(\mathrm{Class}_i \mid \mathrm{Object})$, the object probability Pr(Object), and the predicted intersection over union with the ground truth ($IOU_{pred}^{truth}$) [45]. Due to the aggregation of both class and location confidence, this single value is highly useful as a method of filtering prediction results. The capability to filter predictions by confidence is built into YOLOv7 (and most ML algorithms) by ways of a simple confidence threshold, where only individual predictions above a given value are reported in the final prediction dataset. Here, an individual confidence threshold of 0.4 is used for all analysis, as this value resulted in models with maximum F1 scores. This was selected since high F1 scores correlate to nearly equivalent swelling RMSE values, while the reverse is not necessarily true.

We have further expanded the role of confidence filtering by making two key observations. First, a narrow domain of test images resulted in significantly greater error (with regards to predicted swelling), than most other images. Second, these high error images contained a large amount of low confidence predictions ($\lesssim$0.6) when compared to other images. We hypothesized that while these low confidence predictions were easily filtered out, their overabundance could be indicative of an image that is outside the domain of the trained model. This hypothesis led to the development of a novel image confidence score, in which an overall individual image confidence score was calculated using Equation 7, which weighs both the apparent size of each feature as well as its confidence.

$$Image\ Confidence\ Score = \frac{\sum_{n=1}^{N} Width_n * Height_n * Confidence_n}{\sum_{n=1}^{N} Width_n * Height_n} \qquad (7)$$

This image confidence score is used as an equivalent method to the individual prediction confidence. When a constant threshold filter is applied, only images with an image confidence value above this threshold are reported. Herein, an image confidence threshold of 0.7 is used for all analysis. The use of image confidence provides a simple means to assess model domain and flag potentially problematic images to quantify.

**Domain Expert Labeling Survey.** The labeling survey image set (5 experimental images not used in other training, validation or test, Table 3) was submitted to six early career researchers highly familiar with cavity labeling, *e.g.*, cavity labeling is or has been a significant fraction of their research labor expenditures. The researchers' experience ranged from several years into graduate studies in a relevant field, to early career researchers at a government-sponsored national laboratory. Three of the six have first author publications in which cavity labeling is a key analytical technique (their citations are hidden to maintain anonymity in the results). These researchers learned electron microscopy and data labeling techniques from four different research groups at three major universities, ensuring a variety of methodologies. Identical instructions and labeling tools were submitted to each expert.

## Data availability

The training data used to prepare both the Experimental and Synthetic models, as well as the common test set of 20 Experimental cavity images is provided on Figshare (https://figshare.com/s/b179e8992064286c2e7c). We also provide the trained weights from the "Best" epochs for both model types, described in the body of this work.

## Code availability

A version of the full software (synthetic data generation) is under continuous development, and thus the version used for this manuscript, or the current version are available through the SIGMA-ML GitHub repository at https://github.com/nomelabs/SIGMA-ML .

## Acknowledgements

This work was funded by the Electric Power Research Institute (EPRI) under award numbers 10012138 and 10012245. Additional support for M.J. Lynch was funded by the NASA Space Technology Graduate Research Opportunities (NSTGRO). M.J. Lynch and K.G. Field would like to acknowledge critical feedback from research group members on content and visuals used within this manuscript.

## Author contributions

M.J.L. (conceptualization, data generation & analysis, paper writing, funding acquisition), R.J. (assisted in data analysis, paper editing), G.B. (assisted in data collection, paper editing), P.P. (assisted in machine learning training), D.M. (conceptualization, project supervision, paper editing, funding acquisition), K.G.F. (conceptualization, project supervision, paper editing, funding acquisition).

## Competing Interests

The authors declare no competing interests.

## Corresponding authors

Correspondence to Matthew J. Lynch.